# State Transition Modeling of the Smoking Behavior using LSTM Recurrent Neural Networks

Chrisogonas O. Odhiambo
Computer Science & Engineering
University of South Carolina
Columbia, SC, USA
odhiambo@email.sc.edu

Casey A. Cole
Computer Science & Engineering
University of South Carolina
Columbia, SC, USA
coleca@cse.sc.edu

Alaleh Torkjazi
Computer Science & Engineering
University of South Carolina
Columbia, SC, USA
alale@email.sc.edu

Homayoun Valafar
Computer Science & Engineering
University of South Carolina
Columbia, SC, USA
homayoun@cse.sc.edu

**Abstract** – *The use of sensors has pervaded everyday life in several applications including human activity monitoring, healthcare, and social networks. In this study, we focus on the use of smartwatch sensors to recognize smoking activity. More specifically, we have reformulated the previous work in detection of smoking to include in-context recognition of smoking. Our presented reformulation of the smoking gesture as a state-transition model that consists of the mini-gestures hand-to-lip, hand-on-lip, and hand-off-lip, has demonstrated improvement in detection rates nearing 100% using conventional neural networks. In addition, we have begun the utilization of Long-Short-Term Memory (LSTM) neural networks to allow for in-context detection of gestures with accuracy nearing 97%.*

## I. INTRODUCTION

Cigarette smoking has remained the leading preventable cause of death in the world for the past several decades. In the US alone, 20% of the population report that they engage in smoking and diseases caused by smoking cost the population over $170 billion in healthcare each year (www.cdc.gov, www.who.int)[1]. In addition, the majority of smokers report that they want to quit yet among those that make a quit attempt, the majority relapse at least once. Reducing the number of relapses is of great interest to the community of tobacco-related researchers. Many studies[2]–[5] have been conducted in an attempt to both properly describe smoking behavior as well as pinpointing the best-times to intervene such that a relapse does not occur. However, these studies are inherently limited due to the current methods of studying smoking behavior. Most studies conducted rely on participants to self-report their smoking behavior. In various studies[2], [6], [7], the accuracy of self-reporting has been shown to be no more than ~76%. To bypass the reliance on self-reporting, some studies have been conducted in laboratory-based settings. In these studies, participants are required to smoke while being recorded. In addition, some studies[2] also enforce that the participants insert their cigarette into a device that measures attributes, such as puff duration and the interval between puffs, as they smoke. These measures are extremely useful to researchers who study topics like craving and the effects of nicotine withdrawal. Whereas these studies bypass the limitation of self-reporting, participants often report that they felt uncomfortable in the lab environment or dissatisfaction of the smoking experience due to the incorporation of the measurement device.

A potential solution to these limitations is the use of smartwatch devices. The use of smartwatch technology allows the study to be conducted in a smoker's natural environment, therefore eliminating the biases introduced in laboratory settings. In addition, commercial smartwatch devices come with a rich array of sensors that can be utilized, in conjunction with ML techniques, in order to detect a variety of human activities[8]–[10]. Automatic detection of behaviors allows for an unobtrusive and passive collection and characterization of human activities that does not rely on self-reporting. It also allows for "in-time" intervention techniques to be developed. Our previous work[6] in this field has indicated that smoking can be detected using only accelerometer data and single-layer artificial neural networks (ANNs) with an accuracy of ~95% in laboratory settings and ~90% in real-world settings[6], [11]. Independent reports[12]–[16] also confirmed the usability of smartwatches in the study of human behavior.

Interpretation of human activity can substantially benefit from in-context analysis since there exist temporal relationship between activities. For instance, one smoking gesture clearly consists of a sequence of three consecutive mini-gestures initiated by hand-to-lip, followed by a duration of hand-on-lip, and concluded by a mini-gesture of hand-off-lip. To characterize smoking at this more fine-grain level of mini-gestures will require the reformulation of the classification problem while providing the benefit of improved gesture detection. The primary focus of the presented research is to investigate the impact of mini-gesture representation of smoking. To that end we will explore the performance of conventional and LSTM Neural Networks[17] and compare the results to the previously published work.

## II. BACKGROUND AND METHOD
### A. Previous and Related Work

This work was funded by NIH grant number P20 RR-016461

Considering their rich array of sensors, the cost, accessibility, and ease of use, smartwatches have emerged as a compelling platform to study human activities unobtrusively. Smartwatches have been used as step-counters[16], sleep monitoring[18], diet monitoring[13] as well as general fitness tracking[19]. In the context of smoking, smartwatches have demonstrated to be usable for in-situ study of smoking[20], [21] with high accuracy[6], [20]–[22]. Smartwatches have been used to detect smoking gesture with 95% accuracy in laboratory environment[23] and 90% in-situ detection of smoking[6]. Study of smoking has also been demonstrated to be more accurate when compared to self-reporting (90% versus 78%)[6], [24].

In this experiment we have utilized the smoking data recorded from five smokers and compare our results to the previously published work.[cite] The previously reported detection of smoking was implemented using a hybrid approach that consisted of an ANN for low level detection of smoking puffs alongside a rule based AI for overall classification of detected puffs into smoking sessions. Using this model, a session level detection accuracy of 81-90% was achieved. Whereas this demonstrated a high degree of success, improvements to the detection of individual puffs has the potential to increase this accuracy even further.

### B. Data Annotation

In this experiment we used data from a previously published work that in available for download from https://ifestos.cse.sc.edu. This data consisted of 10 smoking events along with 15 non-smoking events collected across six individuals. Five of the individuals collected their data in a laboratory setting and one in real world settings. The non-smoking activities collected ranging from eating/drinking to typing on the computer. In total, 172 individual smoking puffs were collected. An example of a smoking puff is shown in Figure 1.

In the previous study, each smoking puff was annotated by indicating the start and end of each puff by an expert. Using Matlab, each puff gesture was further parsed into three sub-gestures: hand-to-lip, hand-on-lip, and hand-off-lip. The hand-to-lip gesture was defined as the motion of the cigarette from the resting position (hip, thigh, etc.) to the mouth. This region is shown in the left most box in Figure 2 and encompasses about 20 data points (or 0.8 seconds). The hand-off-lip gesture is then defined as the return of the participant's hand to a resting place. This region is shown in as the far-right box in Figure 2 and is also about 0.8 seconds in length. The hand-on-lip gesture was defined to be the region in which the person has the cigarette in or near their mouth (inhalation time) and is shown in between the two boxes in Figure 2. This region is typically longer than both the hand-to-lip and hand-off-lip regions and greatly varied across participants as well as within each participant. Each hand-on-lip gesture was further broken apart using a rolling window of size 20 to make them compatible with the other mini gestures. Non-smoking gestures were also extracted in this way.

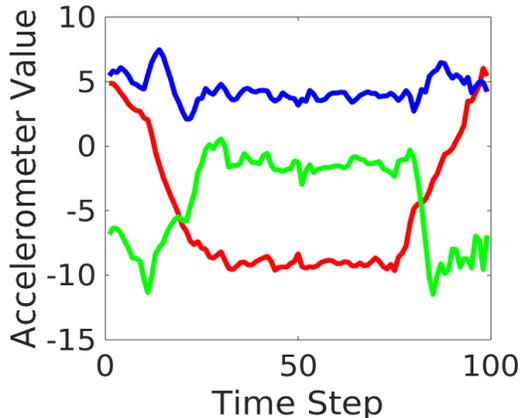

Figure 1. An example of a smoking puff is shown where red indicates the X dimension of the accelerometer data, green the Y and blue the Z.

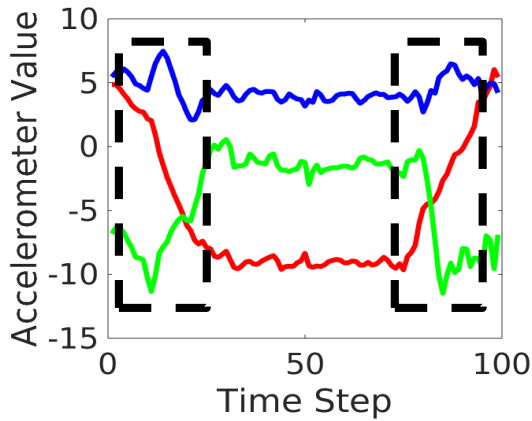

Figure 2. An example of sub-gesture annotation of a puff where the first box denotes the hand-to-lip gesture and the second box the hand-off-lip gesture.

### C. Overview of Training/Validation/Testing Sets

The total number of gestures extracted was 172 hand-to-lip, 5054 hand-on-lip, 172 hand-off-lip and 5854 non-smoking. Due to the imbalance of data in each class, the hand-to-lip and hand-off-lip sub-gestures were duplicated 30 times to ensure nearly uniform number of observations in each training category. The final total number of gestures per sub-gesture were 5,160 hand-to-lip, 5,054 hand-on-lip, 5,160 hand-off-lip and 5,854 non-smoking. The target classes were coded as shown in Table 1 with non-smoking labeled as 1, hand-off-lip as 4, hand-on-lip as 3 and hand-to-lip as 2. One-hot encoding was used to generate a target matrix (one-hot encoding for each class shown in Table 1).

*Table 1. Class assignment and one-hot encoding for each sub-gesture.*

| Sub-gesture | Class | One-hot Encoding |
|---|---|---|
| Non-Smoking | 1 | 1000 |
| Hand-to-lip | 2 | 0100 |
| Hand-on-lip | 3 | 0010 |
| Hand-off-lip | 4 | 0001 |

The resulting dataset was split into the three traditional datasets of training, cross-validation, and testing sets in the ratios of 70%, 15%, 15%, respectively.

### D. Neural Network Platform and Architecture

Using Keras and Tensorflow as the simulation platform of our ANN, we investigated the performance of the conventional feedforward perceptrons, and LSTM neural networks. Both networks were trained using the same dataset published by a previous study[23]. The following sections provide the details for each individual study. We trained both the conventional Artificial Neural Network and Recurrent Neural Network Long-Short-Term Memory (LSTM) types of networks. While ANN generally registered high accuracies, LSTM comes with unique advantages such as constant error backpropagation within memory cells which makes it possible to bridge very long-time lags. It also works well over a broad range of parameters. Importantly, the network's previous knowledge/output forms the input of the next unit. This means that LSTM's learning becomes better with every subsequent unit.

*Conventional ANN* – As the first step in our investigation, we explored the performance of a conventional (shallow) neural network in order to establish the impact of mini-gesture detection instead of detection of an entire smoking puff. Here we implemented a comparable architecture to the one from our previous study[23] and observed the implications of the reformulated study. In our previous study, we utilized a conventional ANN consisting of 300 input neurons, 10 hidden neurons and a single output neuron. The reformulation of the mini-gesture detection requires 60 input neurons and 4 output neurons representing 4 classes of mini-gestures. In this study we investigated the number of hidden layers and hidden neurons that will provide the optimal detection performance. During our studies, we investigated activation function, number of batches, and the number of layers while using Adam optimization[25] method and Binary-crossentropy loss function for training the network. A brief summary of the Keras python code is shown in Figure 3. This code segment was modified to incorporate 2-lyaer, 3-layer, and 4 layers of hidden neurons.

```
model = Sequential()
model.add(Dense(12, input_dim = 60,
activation='relu'))
model.add(Dense(8, activation
= 'relu')) #... (ann-i)
model.add(Dense(4,activation='sigmoid')
)
model.compile(optimizer='adam',
loss='binary_crossentropy',
metrics=['accuracy'])
```

*Figure 3. A snippet of Python code used in Keras to define the used ANN.*

*LSTM-NN* – While the conventional multi-layered feedforward ANNs remain excellent tools to be used in the prediction and classification tasks, they poorly incorporate temporal information. LSTM recurrent models[17], [26] have demonstrated success in incorporating temporal and historical information to their classification protocol and address a very critical aspect of the reformulated data i.e. the temporal aspect. For instance, a typical and permissible smoking puff should consist of the specific sequence of Hand-To-Lip, followed by Hand-On-Lip, and terminated by a Hand-Off-Lip mini-gestures. Furthermore, a typical puff should consist of approximately a reasonable duration of puff (identified by the Hand-On-Lip mini-gesture) that is no shorter than 0.5 second and no longer than 3 seconds. Any departure from this allowed range should disqualify the identification of a proper puff. All these relationships can define a smoking grammar based on the vocabulary of mini-gestures.

Our preliminary investigation of the LSTM-NN consisted of an exploration over the most optimal architecture (number of units) where the input and output of the network consisted of 60 and 4 neurons respectively. Figure 4 illustrates our model of LSTM in the Keras environment. Using this model, we have investigated the performance of varying 2-unit, 3-unit, and 4-unit LSTM architectures.

```
model = Sequential()
model.add(LSTM(output_leng,
batch_input_shape=(None,1,input_leng),
return_sequences=True,
activation='sigmoid')) #... (lstm-i)
model.compile(loss='mse',

optimizer='adam',metrics=['accuracy'])
```

*Figure 4. A snippet of Python code used in Keras to define the used LSTM.*

## E. Training and Testing Procedure

In our investigations, we used the hold-out strategy to set aside a section of the training dataset as a validation set that constitutes a fully independent data set. This strategy has a lower computational cost compared to k-fold strategy because it is only executed once. However, performance evaluation is subject to higher variance, given the smaller size of the data. The entire data set was partitioned using the ratios of 70:15:15 for training set, validation set, and test set respectively.

We evaluated our conventional and LSTM neural networks in terms of loss and accuracy, as a function of architecture. Accuracy measures the performance of the network, while the loss function helps in optimizing the parameters of the neural networks. The objective of the training is to minimize the loss by optimizing parameters i.e. weights. We calculate loss by matching the annotated target value and the predicted values by the network. We used Mean Squared Error (MSE) loss function to quantify the success of our ANN in predicting the desired outputs. We also relied on accuracy measure as an overall metric of classification success.

The primary objective of our exploration was to discover models of ANN that will outperform the previously reported performance of 95%, using the same data set. It is noteworthy, that although in this exercise we used the same data as before, the problem was reformulated such that the input size was reduced from 300 input neurons to 60, and the output neurons was increased from 1 to 4.

## III. RESULTS AND DISCUSSION

In this section we provide the results of our investigations for the optimal performing architectures for the conventional and LSTM neural networks.

### A. Conventional Neural Network

In total, we examined the performance of more than 20 architectures of ANN in order to select the relative optimal architecture. Table 2 shows the performance outcomes at training and testing phases, under some representative architectural and training parameters (epochs, batches and units). Highlights show best performance configurations. The first entry in this table (with the yellow highlight) indicates the architecture with the most optimal performance based on the training set. The loss and accuracy functions for this network as a function of epochs are shown in Figure 5 and Figure 6 respectively. Careful examination of these figures indicates an overtraining of the network based on the increasing pattern of the loss of the validation set. Based on this observation, we imposed two additional constraints that directly relates to our application. The first criterion was to select the most optimal network based on the minimum combined loss of the training and validation set. The green entries in Table 2 denote all configurations that resulted in a combined loss value of 0.06. The second criterion selects against the larger ANN due to the nature of our application that is cognizant of power consumption. Our final selected architecture is shown in blue that balances detection performance and power consumption. Finally,

*Table 2. A summary of mini-gesture detection using different architectures of a conventional ANN. Highlights show best model configurations.*

| Epoch | Batch | Layers | Loss | Val. loss | Accuracy (%) | % Val. Accuracy | Test (%) |
|---|---|---|---|---|---|---|---|
| 5000 | 100 | 4 | 0.01 | 0.07 | 99.54 | 98.84 | 98.79 |
| 5000 | 50 | 4 | 0.01 | 0.09 | 99.47 | 98.70 | 99.65 |
| 5000 | 50 | 3 | 0.02 | 0.06 | 99.34 | 98.72 | 99.28 |
| 5000 | 100 | 3 | 0.02 | 0.06 | 99.20 | 98.72 | 99.18 |
| 5000 | 100 | 2 | 0.03 | 0.06 | 98.98 | 98.43 | 98.57 |
| 3000 | 50 | 4 | 0.01 | 0.07 | 99.49 | 98.76 | 99.29 |
| 3000 | 100 | 4 | 0.02 | 0.05 | 99.36 | 98.88 | 99.59 |
| 3000 | 50 | 2 | 0.02 | 0.06 | 99.31 | 98.82 | 98.40 |
| 3000 | 50 | 3 | 0.03 | 0.07 | 99.07 | 98.69 | 99.17 |
| 2000 | 50 | 4 | 0.02 | 0.04 | 99.41 | 98.91 | 99.57 |
| 2000 | 50 | 3 | 0.02 | 0.07 | 99.13 | 98.59 | 99.13 |
| 2000 | 50 | 2 | 0.03 | 0.06 | 99.13 | 98.46 | 98.21 |
| 2000 | 100 | 4 | 0.03 | 0.06 | 99.13 | 98.65 | 99.38 |
| 2000 | 100 | 3 | 0.03 | 0.06 | 98.91 | 98.48 | 98.79 |
| 1500 | 50 | 3 | 0.02 | 0.05 | 99.37 | 98.81 | 99.13 |
| 1500 | 50 | 4 | 0.02 | 0.04 | 99.35 | 98.85 | 99.42 |
| 1500 | 100 | 4 | 0.02 | 0.04 | 99.18 | 98.70 | 99.26 |
| 1000 | 50 | 4 | 0.02 | 0.05 | 99.09 | 98.55 | 98.69 |
| 1000 | 100 | 3 | 0.03 | 0.06 | 98.92 | 98.54 | 98.68 |

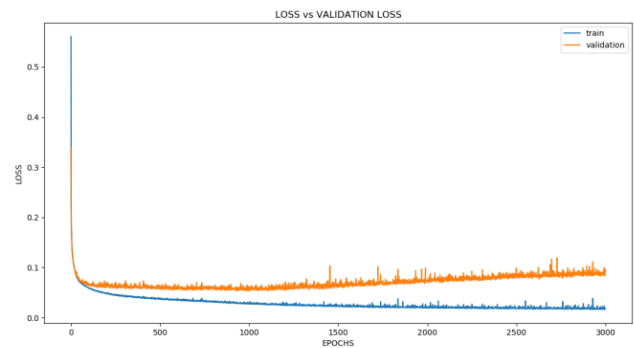

*Figure 5. The loss function of the most optimal ANN architecture as a function of epochs. Training-loss is illustrated in blue and the validation-loss is illustrated in orange.*

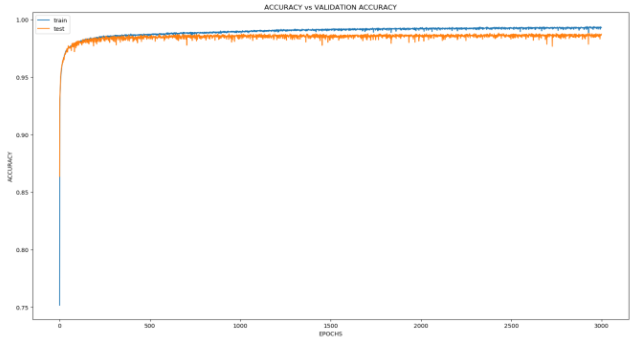

Figure 6. The measure of accuracy as a function of epochs for the training set (blue) and validation set (yellow).

Table 3. The confusion matrix of the most optimal ANN architecture.

| | | Prediction | | | |
|---|---|---|---|---|---|
| | | Rest | H-to-L | H-on-L | H-off-L |
| Actual | Rest | 1125 | 5 | 9 | 3 |
| | H-to-L | 14 | 1021 | 11 | 0 |
| | H-on-L | 59 | 7 | 930 | 9 |
| | H-off-L | 25 | 10 | 0 | 1018 |

Table 4. Confusion matrix report for the most optimal architecture of a conventional ANN.

| | Precision | Recall | f1-score | Support |
|---|---|---|---|---|
| 0 | 0.92 | 0.99 | 0.95 | 1142 |
| 1 | 0.98 | 0.98 | 0.98 | 1046 |
| 2 | 0.98 | 0.93 | 0.95 | 1005 |
| 3 | 0.99 | 0.97 | 0.98 | 1053 |
| micro avg | 0.96 | 0.96 | 0.96 | 4246 |
| macro avg | 0.97 | 0.96 | 0.96 | 4246 |
| weighted avg | 0.97 | 0.96 | 0.96 | 4246 |

### B. LSTM Neural Network

Table 5 provides an overview summary of our investigation of varying LSTM architectures in detection of smoking mini-gestures. Similar to the case of the conventional ANN, our first selections of the optimal architectures are highlighted in yellow. However, motivated by reducing the battery power consumption, we imposed the minimalism of architecture without any compromise of the performance. By maintaining a sum of loss of 0.06 over the training and validation while reducing the number of LSTM units, the LSTM configuration highlighted in blue is selected. As an indirect consequence of this selection, the performance of the network on the test set increased from approximately 93% to 95% indicating slight memorization by the network that can be remedied by smaller LSTM networks. The memorization phenomenon can also be confirmed after careful examination of the loss and accuracy functions of the training and validation sets shown in Figure 7 and Figure 8 respectively. Table 3 illustrates the confusion matrix of the most optimal ANN, while Table 4 provides a summary of the precision and recall of the confusion table.

Table 5. A summary of LSTM's performance in detection smoking mini-gesture as a function of different architectural parameters. Highlights show best model configurations.

| epoch | Batch | Units | Loss | Val. Loss | Accuracy (%) | (%) Val. Accuracy | Test Acc. (%) |
|---|---|---|---|---|---|---|---|
| 5000 | 100 | 3 | 0.03 | 0.03 | 96.90 | 95.74 | 93.57 |
| 5000 | 50 | 3 | 0.03 | 0.03 | 96.00 | 95.29 | 94.85 |
| 5000 | 100 | 4 | 0.03 | 0.03 | 95.92 | 94.77 | 93.46 |
| 5000 | 100 | 2 | 0.03 | 0.04 | 95.54 | 94.25 | 93.00 |
| 5000 | 50 | 4 | 0.05 | 0.05 | 94.56 | 93.74 | 95.51 |
| 3000 | 50 | 3 | 0.03 | 0.03 | 96.67 | 95.48 | 94.78 |
| 3000 | 50 | 2 | 0.03 | 0.03 | 96.21 | 95.20 | 95.02 |
| 2000 | 50 | 2 | 0.04 | 0.04 | 94.85 | 93.95 | 95.31 |
| 2000 | 50 | 3 | 0.04 | 0.04 | 94.80 | 94.14 | 95.72 |
| 2000 | 50 | 4 | 0.04 | 0.04 | 94.65 | 93.90 | 94.97 |
| 2000 | 100 | 4 | 0.05 | 0.05 | 94.61 | 93.88 | 96.13 |
| 2000 | 100 | 2 | 0.04 | 0.04 | 94.60 | 94.11 | 93.39 |
| 1500 | 100 | 3 | 0.03 | 0.04 | 95.95 | 94.94 | 95.13 |
| 1500 | 50 | 3 | 0.04 | 0.04 | 94.67 | 94.23 | 94.92 |
| 1000 | 50 | 3 | 0.03 | 0.03 | 95.51 | 94.98 | 94.44 |
| 1000 | 100 | 4 | 0.05 | 0.05 | 95.26 | 94.30 | 95.23 |
| 500 | 50 | 4 | 0.05 | 0.05 | 94.97 | 94.02 | 94.62 |
| 500 | 50 | 2 | 0.04 | 0.04 | 94.92 | 94.80 | 91.90 |
| 500 | 100 | 3 | 0.04 | 0.04 | 94.47 | 94.21 | 95.01 |

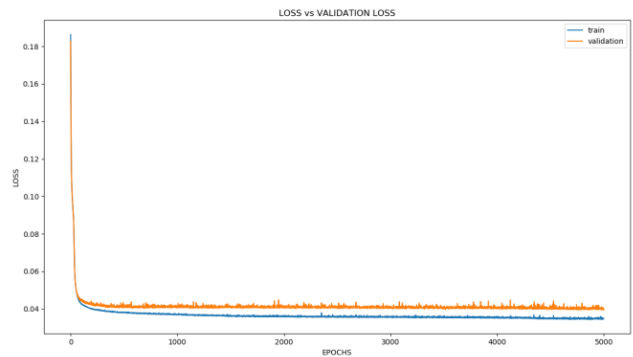

Figure 7. The loss function of the most optimal LSTM architecture as a function of epochs. Training-loss is illustrated in blue and the validation-loss is illustrated in orange.

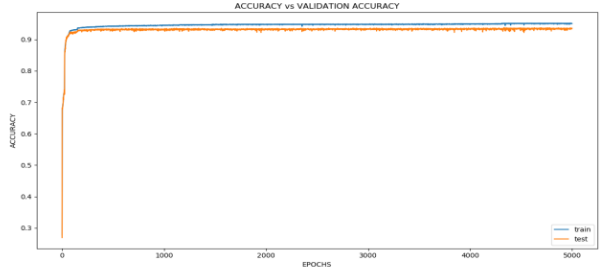

*Figure 8. The measure of accuracy of the most optimal LSTM architecture as a function of epochs for the training (blue) and validation sets (orange).*

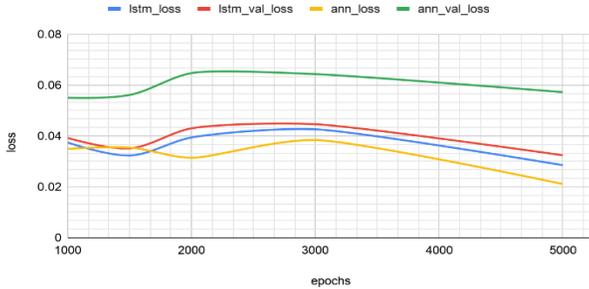

*Figure 9. Comparing LSTM/ANN loss. Blue is LSTM loss, red is LSTM validation loss, light orange is ANN loss, and green is ANN validation loss.*

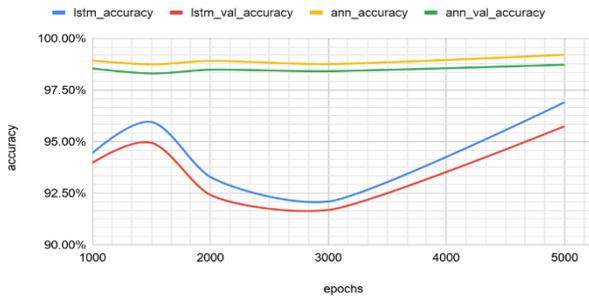

*Figure 10. Comparing LSTM/ANN Accuracy. Blue is LSTM accuracy, red is LSTM validation accuracy, light orange is ANN accuracy, and green is ANN validation accuracy.*

Figure 9 shows ANN validation loss is generally higher than in LSTM for all trained models. Figure 10 shows both ANN validation accuracy and accuracy consistently higher than in LSTM for all trained models.

*Table 6. The confusion matrix of the most optimal LSTM architecture. Batch=100, Units=3, Activation=sigmoid, epochs=5000*

|        |         | Prediction |        |        |         |
|--------|---------|------------|--------|--------|---------|
|        |         | Rest       | H-to-L | H-on-L | H-off-L |
| Actual | Rest    | 1090       | 7      | 33     | 12      |
|        | H-to-L  | 17         | 1017   | 12     | 0       |
|        | H-on-L  | 43         | 23     | 929    | 10      |
|        | H-off-L | 10         | 25     | 24     | 994     |

*Table 7. Confusion matrix report for the most optimal architecture of LSTM.*

|              | Precision | Recall | f1-score | Support |
|--------------|-----------|--------|----------|---------|
| 0            | 0.94      | 0.95   | 0.95     | 1142    |
| 1            | 0.95      | 0.97   | 0.96     | 1046    |
| 2            | 0.93      | 0.92   | 0.93     | 1005    |
| 3            | 0.98      | 0.94   | 0.96     | 1053    |
| micro avg    | 0.95      | 0.95   | 0.95     | 4246    |
| macro avg    | 0.95      | 0.95   | 0.95     | 4246    |
| weighted avg | 0.95      | 0.95   | 0.95     | 4246    |

## IV. CONCLUSION

In this report, we have presented the reformulation of an entire smoking gesture (puff) as a combination of three time-dependent mini-gestures (hand-to-lip, hand-on-lip, and hand-off-lip). Using this reformulation, we demonstrated the success of conventional ANN (99%) in improving upon the previously reported detection of smoking (95%) using the same set of data. Based on the results shown in Table 2, the reformulation of the smoking gesture as mini-gestures clearly reduces the complexity of detection as evidenced by the improved detection. Although we have achieved a near perfect detection of the smoking gesture, we anticipate unforeseen challenges during the live deployment of this technology for in-situ study of human smoking behavior. Furthermore, we remain cognizant of the battery requirement during the live deployment of this technology.

In order to incorporate the temporal dependency of human activities, including the mini-gestures, we have hypothesized that LSTM recurrent neural networks would exhibit a better performance. While our initial and in laboratory investigations have not supported this hypothesis, we anticipate that the true value of RNN will be exposed in live deployment of the system.

In summary, our state-transition approach to detection of smoking mini-gestures had demonstrated improvements over the previously reported approaches. We expect that the declaration of mini-gestures as the "vocabulary" of smoking is instrumental in the development of the smoking "grammar" that can be exploited by the incorporation of RNNs.

**REFERENCE**
[1] K. E. Warner, T. A. Hodgson, and C. E. Carroll, "Medical costs of smoking in the United States: estimates, their validity, and their implications," *Tob. Control*, vol. 8, pp. 290–300, 1999.
[2] S. De Jesus, A. Hsin, G. Faulkner, and H. Prapavessis, "A systematic review and analysis of data reduction

techniques for the CReSS smoking topography device," *J. Smok. Cessat.*, vol. 10, no. 1, pp. 12–28, Oct. 2013.
[3] E. T. Moolchan, C. S. Parzynski, M. Jaszyna-Gasior, C. C. Collins, M. K. Leff, and D. L. Zimmerman, "A link between adolescent nicotine metabolism and smoking topography," *Cancer Epidemiol. Biomarkers Prev.*, vol. 18, no. 5, pp. 1578–1583, May 2009.
[4] F. H. Franken, W. B. Pickworth, D. H. Epstein, and E. T. Moolchan, "Smoking rates and topography predict adolescent smoking cessation following treatment with nicotine replacement therapy," *Cancer Epidemiol. Biomarkers Prev.*, vol. 15, no. 1, pp. 154–157, Jan. 2006.
[5] L. W. Frederiksen, P. M. Miller, and G. L. Peterson, "Topographical components of smoking behavior," *Addict. Behav.*, vol. 2, no. 1, pp. 55–61, 1977.
[6] C. A. Cole, D. Anshari, V. Lambert, J. F. Thrasher, and H. Valafar, "Detecting Smoking Events Using Accelerometer Data Collected Via Smartwatch Technology: Validation Study.," *JMIR mHealth uHealth*, vol. 5, no. 12, p. e189, Dec. 2017.
[7] D. L. Patrick, A. Cheadle, D. C. Thompson, P. Diehr, T. Koepsell, and S. Kinne, "The validity of self-reported smoking: A review and meta-analysis," *Am. J. Public Health*, vol. 84, no. 7, pp. 1086–1093, 1994.
[8] M. Ashfak Habib, M. S. Mohktar, S. Bahyah Kamaruzzaman, K. Seang Lim, T. Maw Pin, and F. Ibrahim, "Smartphone-based solutions for fall detection and prevention: Challenges and open issues," *Sensors (Switzerland)*, vol. 14, no. 4. MDPI AG, pp. 7181–7208, 22-Apr-2014.
[9] "SmartFall: A Smartwatch-Based Fall Detection System Using Deep Learning." [Online]. Available: https://www.ncbi.nlm.nih.gov/pmc/articles/PMC6210545/. [Accessed: 08-Nov-2019].
[10] C. Ma, D. Wong, W. Lam, A. Wan, and W. Lee, "Balance Improvement Effects of Biofeedback Systems with State-of-the-Art Wearable Sensors: A Systematic Review," *Sensors*, vol. 16, no. 4, p. 434, Mar. 2016.
[11] C. A. Cole, J. F. Thrasher, S. M. Strayer, and H. Valafar, "Resolving ambiguities in accelerometer data due to location of sensor on wrist in application to detection of smoking gesture," in *2017 IEEE EMBS International Conference on Biomedical and Health Informatics, BHI 2017*, 2017, pp. 489–492.
[12] M. Shoaib, S. Bosch, O. Incel, H. Scholten, and P. Havinga, "Complex Human Activity Recognition Using Smartphone and Wrist-Worn Motion Sensors," *Sensors*, vol. 16, no. 4, p. 426, Mar. 2016.
[13] S. Sen, V. Subbaraju, A. Misra, R. K. Balan, and Y. Lee, "The case for smartwatch-based diet monitoring," in *2015 IEEE International Conference on Pervasive Computing and Communication Workshops, PerCom Workshops 2015*, 2015, pp. 585–590.
[14] G. M. Weiss, J. L. Timko, C. M. Gallagher, K. Yoneda, and A. J. Schreiber, "Smartwatch-based activity recognition: A machine learning approach," in *3rd IEEE EMBS International Conference on Biomedical and Health Informatics, BHI 2016*, 2016, pp. 426–429.
[15] T. R. Mauldin, M. E. Canby, V. Metsis, A. H. H. Ngu, and C. C. Rivera, "Smartfall: A smartwatch-based fall detection system using deep learning," *Sensors (Switzerland)*, vol. 18, no. 10, Oct. 2018.
[16] V. Genovese, A. Mannini, and A. M. Sabatini, "A Smartwatch Step Counter for Slow and Intermittent Ambulation," *IEEE Access*, vol. 5, pp. 13028–13037, 2017.
[17] D. Arifoglu and A. Bouchachia, "Activity Recognition and Abnormal Behaviour Detection with Recurrent Neural Networks," in *Procedia Computer Science*, 2017, vol. 110, pp. 86–93.
[18] X. Sun, L. Qiu, Y. Wu, Y. Tang, and G. Cao, "SleepMonitor," *Proc. ACM Interactive, Mobile, Wearable Ubiquitous Technol.*, vol. 1, no. 3, pp. 1–22, Sep. 2017.
[19] A. Pfannenstiel and B. S. Chaparro, "An investigation of the usability and desirability of health and fitness-tracking devices," in *Communications in Computer and Information Science*, 2015, vol. 529, pp. 473–477.
[20] N. Saleheen *et al.*, "PuffMarker: A multi-sensor approach for pinpointing the timing of first lapse in smoking cessation," in *UbiComp 2015 - Proceedings of the 2015 ACM International Joint Conference on Pervasive and Ubiquitous Computing*, 2015, pp. 999–1010.
[21] A. L. Skinner, C. J. Stone, H. Doughty, and M. R. Munafò, "StopWatch:The preliminary evaluation of a smartwatch-based system for passive detection of cigarette smoking," *Nicotine and Tobacco Research*, vol. 21, no. 2. Oxford University Press, pp. 257–261, 01-Feb-2019.
[22] C. A. Cole, D. Anshari, V. Lambert, J. F. Thrasher, and H. Valafar, "Detecting Smoking Events Using Accelerometer Data Collected Via Smartwatch Technology: Validation Study," *JMIR mHealth uHealth*, vol. 5, no. 12, p. e189, Dec. 2017.
[23] "(16) (PDF) Recognition of Smoking Gesture Using Smart Watch Technology." [Online]. Available: https://www.researchgate.net/publication/315720056_Recognition_of_Smoking_Gesture_Using_Smart_Watch_Technology. [Accessed: 08-Nov-2019].
[24] L. E. Wagenknecht, G. L. Burke, L. L. Perkins, N. J. Haley, and G. D. Friedman, "Misclassification of smoking status in the CARDIA study: A comparison of self-report with serum cotinine levels," *Am. J. Public Health*, vol. 82, no. 1, pp. 33–36, 1992.
[25] D. P. Kingma and J. Lei Ba, "ADAM: A METHOD FOR STOCHASTIC OPTIMIZATION."
[26] J. C. Núñez, R. Cabido, J. F. Vélez, A. S. Montemayor, and J. J. Pantrigo, "Multiview 3D human pose estimation using improved least-squares and LSTM networks," *Neurocomputing*, vol. 323, pp. 335–343, Jan. 2019.